\documentclass[final,5p,twocolumn, compress]{elsarticle}

\makeatletter
\def\ps@pprintTitle{%
 \let\@oddhead\@empty
 \let\@evenhead\@empty
 \let\@oddfoot\@empty
 \let\@evenfoot\@empty}
\makeatother

\usepackage{amssymb}
\usepackage{amsmath}
\usepackage{latexsym}

\usepackage{enumitem}

\usepackage{color}
\usepackage{times}
\usepackage{soul}

\usepackage{multirow}

\usepackage{amsthm}
\usepackage{booktabs}

\usepackage[switch]{lineno}
\usepackage{CJKutf8}
\usepackage{subcaption}
\usepackage{bbding}
\usepackage{tabularx}
\usepackage{apalike}

\usepackage{titlesec}
\usepackage[colorlinks,
            linkcolor=red,
            anchorcolor=blue,
            citecolor=black
            ]{hyperref}
\AtBeginDocument{
  \hypersetup{
    linkcolor=red,      
    citecolor=black,    
    filecolor=mycustompurple, 
    urlcolor=magenta       
  }
}

\journal{}
\biboptions{authoryear}
\begin{document}

\begin{frontmatter}

\title{Balanced Residual Distillation Learning for 3D Point Cloud Class-Incremental Semantic Segmentation} 

\author[1]{Yuanzhi Su}
\ead{yuanzhisu@e.gzhu.edu.cn}
\author[1]{Siyuan Chen\corref{mycorrespondingauthor}}
\ead{chensiyuan@gzhu.edu.cn}
\author[1]{Yuan-Gen Wang\corref{mycorrespondingauthor}}
\cortext[mycorrespondingauthor]{Corresponding authors.}
\ead{wangyg@gzhu.edu.cn}
\affiliation[1]{organization={School of Computer Science and Cyber Engineering},
            addressline={Guangzhou University}, 
            city={Guangzhou},
            state={Guangdong},
            country={China}}

\begin{abstract}
Class-incremental learning (CIL) enables continuous learning of new classes while mitigating catastrophic forgetting of old ones. For the performance breakthrough of CIL, it is essential yet challenging to effectively refine past knowledge from the base model and balance it with new learning. However, such a challenge has not been considered in current research. This work proposes a balanced residual distillation learning framework (BRDL) to address this gap and advance CIL performance. BRDL introduces a residual distillation strategy to dynamically refine past knowledge by expanding the network structure and a balanced pseudo-label learning strategy to mitigate class bias and balance learning between old and new classes. We apply the proposed BRDL to a challenging 3D point cloud semantic segmentation task where the data is unordered and unstructured. Extensive experimental results demonstrate that BRDL sets a new benchmark with an outstanding balance capability in class-biased scenarios.
\end{abstract}


\end{frontmatter}

\section{Introduction}
\label{sec: intro}
\begin{figure}[t]
\centering
\includegraphics[width=\linewidth,scale=0.001]{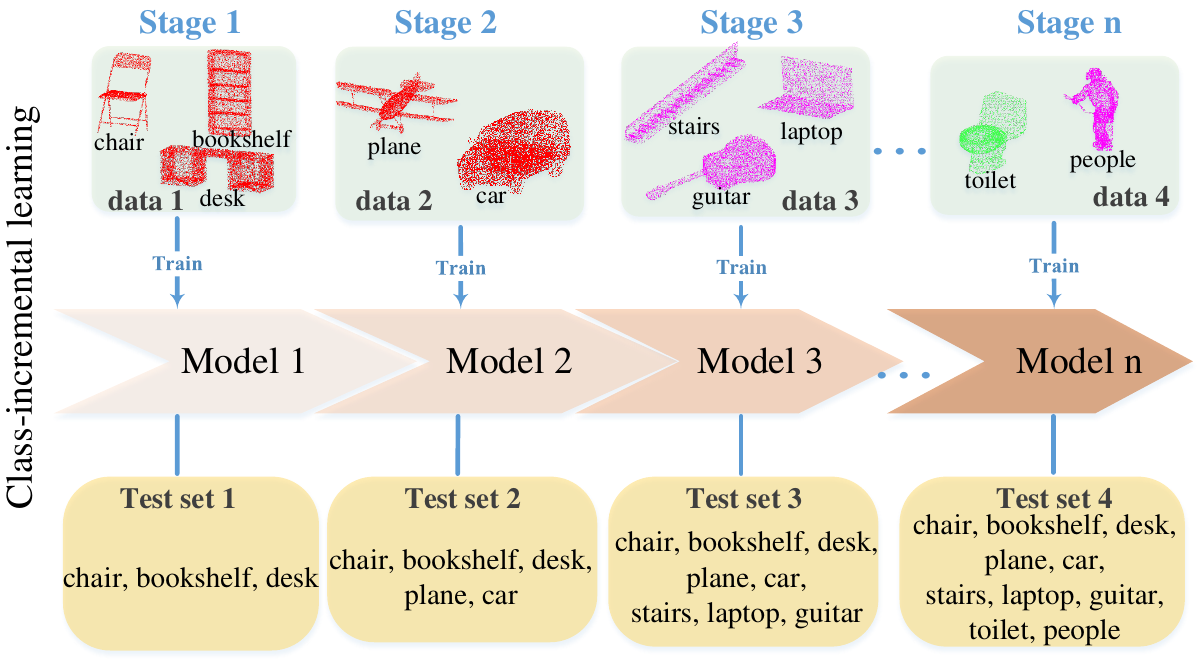}
\caption{Illustration of class-incremental learning process. In each stage, the model is incrementally trained on a new set of classes while retaining knowledge of previously learned ones. For instance, at the initial stage, the model is introduced to classes such as ``chair,'' ``bookshelf,'' and ``desk.'' In the subsequent stages, the model learns additional classes, including ``plane'' and ``car,'' thereby expanding its knowledge while preserving its understanding of earlier classes. This iterative process continues across stages, with new classes introduced at each step, allowing the model to consistently integrate new information while preventing the forgetting of previously learned classes.}
\label{intro_cil}
\vspace{2mm}
\end{figure}

The rapid advancement of LiDAR technology has captured significant attention and revolutionized point cloud-based vision tasks, achieving unprecedented success across various real-world applications. From virtual reality~\cite{blanc2020genuage} to autonomous driving systems~\cite{yue2018lidar}, LiDAR's advancements have been instrumental in tackling complex challenges. As a key to scene understanding, 3D point cloud semantic segmentation, which aims to recognize various objects within a scene, has been significantly developed ~\cite{qi2017pointnet,wang2019dynamic,lai2022stratified, xu2023hierarchical, zeng2024large}. In practical scenarios, the continuous update of hardware and software leads to the discovery of previously unknown categories, necessitating the adaptation of neural networks to recognize new classes. The prevailing approach involves retraining models by integrating all available data, thereby enabling them to handle all categories. However, such a strategy encounters two formidable challenges: the substantial computational costs associated with frequent retraining and the uncertainty regarding the availability and quality of previous data. In response to these challenges, class-incremental learning (CIL) emerges as a promising paradigm. By enabling models to progressively assimilate new information while retaining memory of previously acquired knowledge (as depicted in Figure~\ref{intro_cil}), CIL offers an efficient and effective solution that not only alleviates the computational burden of repeated retraining but also enhances the adaptability of models to dynamically evolving environments.

Most of the existing works on CIL concentrate on 2D image classification~\cite{kirkpatrick2017overcoming,li2017learning,rebuffi2017icarl,yan2021dynamically,wang2022foster,zhang2023open, malepathirana2023napa, gao2024apm}, with further efforts made towards 2D image semantic segmentation~\cite{yu2022self,yang2022uncertainty,xiao2023endpoints}. These approaches typically employ regularization~\cite{kirkpatrick2017overcoming,ebrahimi2018uncertainty, fischer2024inemo}, replay~\cite{rebuffi2017icarl,castro2018end,rolnick2019experience, nguyen2024class}, or parameter isolation~\cite{yan2021dynamically,wang2022foster} to retain previously acquired knowledge. However, research on incremental learning for 3D point clouds is very limited. At present, only a few works~\cite{dong2021i3dol,liu2021l3doc,yang2023label} have applied incremental learning to point clouds, simply extending 2D methods to 3D point clouds. Due to the unordered and unstructured nature of 3D point clouds, preserving previous knowledge is challenging, which can lead to catastrophic forgetting~\cite{kirkpatrick2017overcoming} after the incremental phase. Yang et al.~\cite{yang2023geometry} pioneeringly proposed a class-incremental method on 3D point cloud semantic segmentation, which contains a Geometry-aware Feature relation Transfer module (GFT) and an Uncertainty-aware Pseudo-label Generation strategy (UPG). Although Yang et al.'s method improved overall performance, it imposed a strong constraint on acquired parameters during incremental training, which excessively prioritized preserving knowledge from the base model while hindering the learning of new classes. Furthermore, in cases where the volume of new classes is much less than that of the old ones, this imbalance results in skewed class learning and significant performance disparity for new classes compared to joint training.

To solve these issues, this paper presents a balanced residual distillation learning framework (BRDL) for class-incremental semantic segmentation on 3D point clouds, which includes two core designs: preventing the forgetting of acquired knowledge and facilitating the model's adaptation to new classes. Specifically, to tackle the catastrophic forgetting issue associated with unstructured and unordered point clouds, we design a residual distillation learning strategy (RDL), which focuses on learning the differences or residuals between the old and new classes. RDL allows the model to selectively update its parameters based on the characteristics of the new data without completely overwriting previously learned information. This simple yet effective strategy not only enhances inter-class discriminative capability but also reduces constraints on the model, promoting the learning of new classes. Moreover, to address the model's preference for base classes, we design a balanced pseudo-label learning (BPL) strategy. This strategy balances the contribution of old and new classes during the incremental phase by generating a guidance mask based on the similarity between the representations of new samples and the prototypes of base classes. Lastly, for long-term learning without adding other burdens, we introduce structural reparameterization to enable the model to consistently learn from new classes. The major contributions of our work can be summarized as: 
\begin{itemize} 
\setlength{\itemsep}{0pt} 
\setlength{\parsep}{0pt} \setlength{\parskip}{0pt} 
    \item We propose a novel framework BRDL for 3D point cloud class-incremental learning, which jointly protects prior knowledge and acquires new knowledge in CIL. 
    \item We design a residual distillation learning strategy to refine past knowledge and effectively assimilate information from new classes. Additionally, we implement a balanced pseudo-label learning strategy to mitigate the model's bias toward base classes, favoring new classes. 
    \item Extensive experiments demonstrate that our BRDL significantly outperforms existing methods and exhibits outstanding balancing capability. 
\end{itemize}

\section{Related Work}
\subsection{Class-Incremental Learning} 
Class-incremental learning has garnered increasing attention in machine learning, as it requires models to adapt to new classes without forgetting previously learned ones. Various approaches have been developed to address this challenge, including regularization-based, replay-based, distillation-based, and more sophisticated structure-based methods. \\
\textbf{Regularization-based} methods~\cite{kirkpatrick2017overcoming,ahn2019uncertainty} address the issue of catastrophic forgetting by measuring the importance of model parameters and imposing various constraints on them during the incremental phase. Specifically, regularization-based methods assess the importance of different model parameters and assign varying weights to signify their significance. During incremental steps, less critical parameters can be updated more liberally compared to their more pivotal counterparts. Recently, \cite{fischer2024inemo} introduces an incremental neural mesh model and a positional regularization term to allocate feature space for unseen classes, improving performance in both in-domain and out-of-distribution scenarios.\\ 
\textbf{Replay-based} methods~\cite{rebuffi2017icarl,castro2018end,rolnick2019experience} store a small number of exemplars or employ a generative model to synthesize information related to the old classes. ICaRL~\cite{rebuffi2017icarl} initially proposes the nearest-mean-of-exemplars selection strategy to identify the most representative exemplars for each old class. Alternatively, pseudo-rehearsal techniques use generative models to create pseudo-exemplars based on estimated distributions of previous classes. An advanced exemplar~\cite{nguyen2024class} introduces Causal Relational Replay, a novel framework for class-incremental learning that leverages causal reasoning to learn and rehearse intrinsic relations between classes over time. These methods successfully retain the discriminative features of previously learned classes during the incremental steps.\\ 
\textbf{Distillation-based} methods~\cite{zhao2020maintaining,yang2023label} play a vital role in preserving previously acquired knowledge as the model evolves through current or future incremental steps. Additionally, distillation-based methods involve reducing the divergence between the output probability distributions of previous and current models, thereby consolidating the shared information and enhancing the model's overall robustness and adaptability.\\
\textbf{Structure-based} methods~\cite{yoon2018lifelong,hung2019compacting,wang2022foster} select and expand different sub-network structures involved in the optimization process of the incremental steps, which enhance the model's adaptability for accommodating new classes. Nevertheless, traditional approaches to dynamic architecture may result in notable overhead due to the continuous addition of new modules during incremental steps. Recently, \cite{zhou2024expandable} introduced EASE for class-incremental learning using pre-trained models that employ expandable subspaces and semantic-guided prototype complements to learn new classes without forgetting old ones.

Despite the advanced methods~\cite{nguyen2024class, zhou2024expandable} in the 2D image, directly transferring these methods to 3D point clouds faces several challenges, including their inability to handle the sparse, irregular structure of 3D point clouds, and the difficulty in modeling the dynamic, high-dimensional spatial relationships that are essential for 3D point cloud analysis. Notably, a recent method LGKD~\cite{yang2023label} introduces label-guided knowledge distillation to mitigate catastrophic forgetting, thereby enhancing long-term memory and adaptability in continuous learning tasks. However, it's worth noting that LGKD specifically addresses background shifts and can be classified as a distillation-based method. In contrast, our proposal introduces a novel framework that integrates replay, distillation, and structural adaptation. This combination is necessary because replay retains critical exemplars to prevent forgetting, distillation ensures the smooth transfer of knowledge by aligning old and new model representations, and structural adaptation dynamically expands the network to effectively accommodate new classes. The integration of these paradigms addresses the core challenges of class-incremental learning, providing a balanced solution that effectively retains past knowledge while incorporating new information.

\subsection{3D Point Clouds Semantic Segmentation}
3D point cloud semantic segmentation can be viewed as a point-wise classification task that necessitates consideration of point-to-point disparities and topological distributions. Traditional 3D point cloud semantic segmentation methods~\cite{qi2017pointnet,wang2019dynamic, cheng2024edgcnet, wu2024appfnet, zeng2024large} are generally given by a static set of predefined classes, making them inadequate for scenarios where new classes must be incrementally integrated into the segmentation model. Class-incremental learning is proposed to address this challenge. Previous research has explored various techniques for incremental learning in pixel-level segmentation tasks~\cite{kang2022class,malepathirana2023napa}, but directly extending these methods to 3D point clouds presents unique challenges. To our knowledge, only one pioneering work~\cite{yang2023geometry} addresses class-incremental semantic segmentation on 3D point clouds. Although their work achieved state-of-the-art performance in 3D class-incremental semantic segmentation, it still suffers from critical drawbacks, as analyzed in Sec.~\ref{sec: intro}. These limitations motivate our research to explore novel and improved methods for class-incremental semantic segmentation within the context of 3D point clouds.

\subsection{Incremental Learning in Computer Vision}
In recent years, incremental learning has undergone significant expansion, particularly within computer vision tasks. This paradigm shift has led to groundbreaking applications across a wide spectrum of vision tasks, ranging from traditional image classification to more complex challenges like instance segmentation and semantic segmentation. Incremental learning has notably been successfully applied to a diverse array of vision tasks, including but not limited to image classification~\cite{douillard2020podnet,rebuffi2017icarl,ahn2021ss, gao2024apm}, action recognition~\cite{li2021else,park2021class}, instance segmentation~\cite{nguyen2022ifs}, semantic segmentation~\cite{yang2023geometry,malepathirana2023napa}, object detection~\cite{feng2022overcoming,yang2022continual,yin2022sylph, bai2024revisiting}, and self-supervised representation learning~\cite{jang2021towards,madaan2021representational,fini2022self, zhang2024semi}. Building upon this rich research landscape, this paper aims to delve into the complexities of class-incremental learning for 3D point clouds. We seek to address challenges like catastrophic forgetting and the inherent difficulty of learning new classes in dynamic environments. To this end, we contribute advanced techniques for handling incremental learning in the context of 3D point clouds, ultimately enhancing the robustness and adaptability of computer vision systems in real-world applications.

\begin{figure*}[t]   
	\centering
	\includegraphics[width=\linewidth]{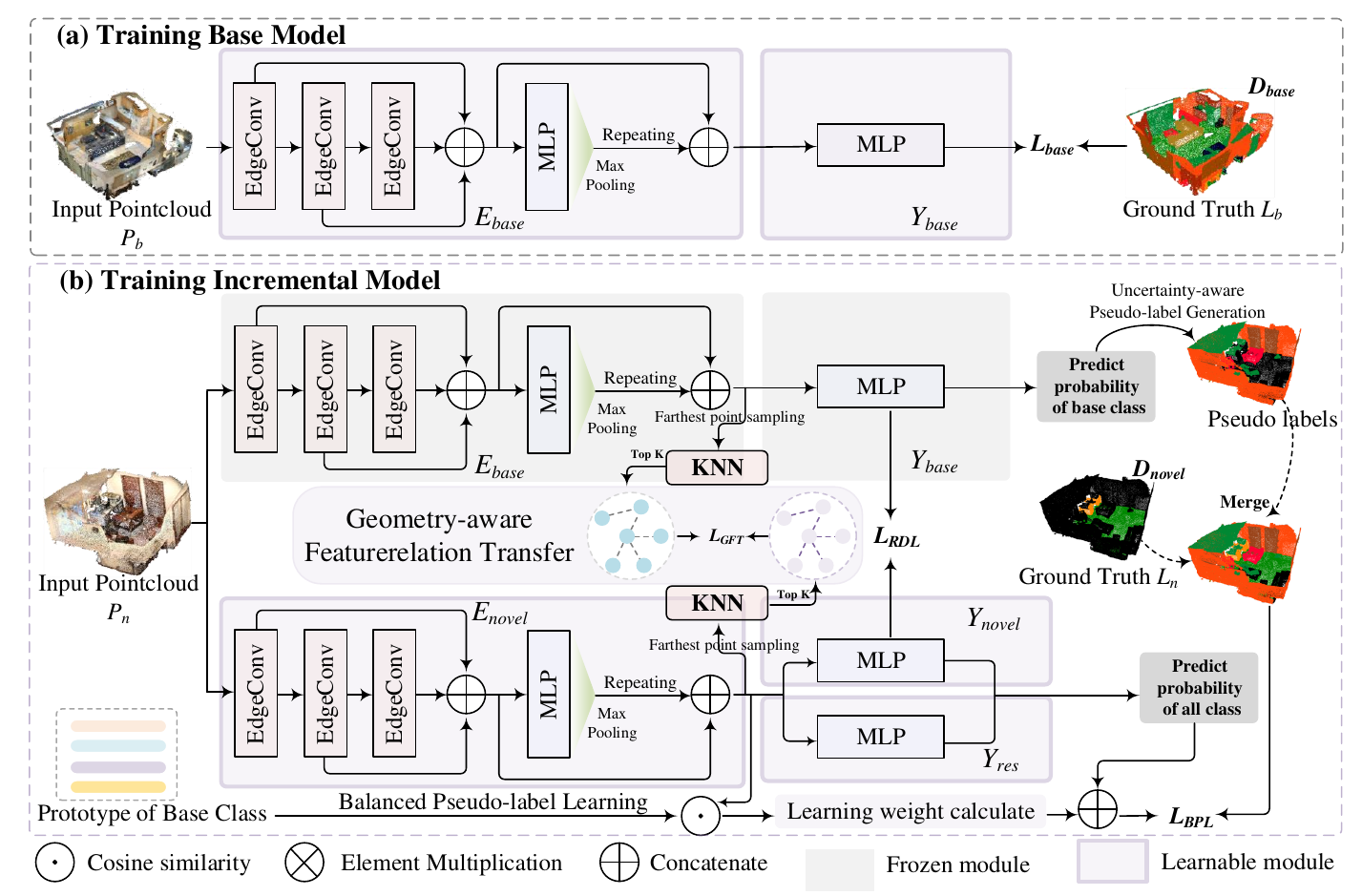}
	\caption{The overall architecture of BRDL. In phase (a), we train the base model consisting of encoder $E_{base}$ and classifier $Y_{base}$ on $D_{base}$. In phase (b), to update the model to segment more classes, we design a residual distillation learning strategy to capture knowledge learned from the base model. Meanwhile, we design a balanced pseudo-label learning strategy to promote the learning of new classes, thereby addressing the training bias.}
	\label{fig: network}
\end{figure*}

\section{Method}
\subsection{Problem Formulation}
Denote the class set and the dataset as $C$ and $D$, respectively. In general, the dataset $D$ collects lots of samples, i.e. scenes. Each scene contains many points, and each point is represented by a pair $(P, L)$. $P \in R^{3+f}$ denotes the input 3D point with $xyz$ coordinates and $f$-dimensional features (e.g., \ RGB), and $L$ denotes the label of the corresponding point. The process of class incremental learning typically includes two phases. In the first phase, the base model $M_{base}$ is trained on the dataset $D_{base}$. In the second phase, $M_{base}$ serves as the initialization for the novel model $M_{novel}$, which is then trained on the dataset $D_{novel}$.
Note that $D_{base}$ is not available during the second phase and $C_{base}$ is disjoint with $C_{novel}$ (i.e., $C_{base} \cap C_{novel} = \varnothing$). Finally, the trained $M_{novel}$ is utilized to perform the segmentation task on both $C_{base}$ and $C_{novel}$.

\begin{figure*}[t]
	\centering
	\includegraphics[width=18cm]{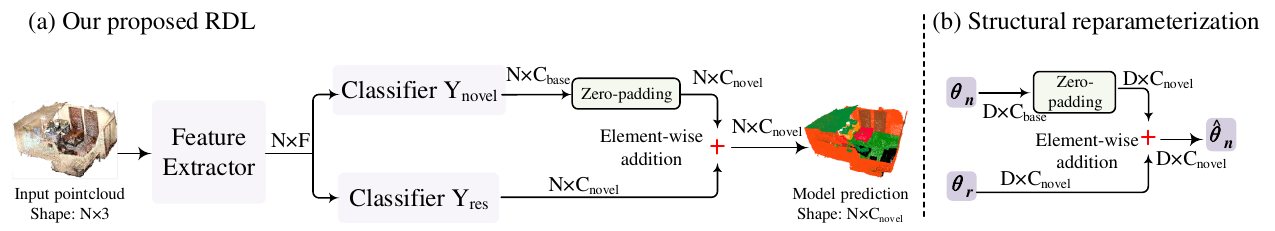}
	\caption{Illustration of our proposed RDL strategy and structural reparameterization.}
	\label{fig:rdl}
\end{figure*}

The core problem of class-incremental learning is protecting existing knowledge and acquiring new knowledge. To this aim, we present a BRDL framework for the CIL task, as shown in Figure~\ref{fig: network}. BRDL consists of Residual Distillation Learning (\textit{cf.} Sec.~\ref{rdl}) and Balanced Pseudo-label Learning (\textit{cf.} Sec.~\ref{bpl}). In the following, we illustrate the above designs in detail.
 
\subsection{Residual Distillation Learning}
\label{rdl}
The notorious catastrophic forgetting in incremental learning is a serious problem. Finding methods to prevent the forgetting of old knowledge while assimilating new information continues to present a challenge. To this aim, we design an RDL strategy to refine the knowledge learned from the $M_{base}$ while effectively absorbing new information, as illustrated in Figure~\ref{fig:rdl}(a). Denote $P_n\in D_{novel}$ as the current input point cloud of novel classes with $S$ points. 

We start by uniformly sampling $r$-proportional points from the input point cloud using farthest point sampling (FPS)~\cite{qi2017pointnet++}. FPS begins by randomly selecting an initial anchor point. Subsequent anchor points are iteratively chosen to maximize their minimum distance from all previously selected anchors. This approach guarantees that the selected anchors are evenly distributed throughout the input point cloud, effectively capturing its overall structure. After the FPS algorithm selects $Z = \lfloor r \times S\rfloor$ anchors, the $l_2$ distance is then used to find the $K$ nearest points, creating local geometric structures. This enables capturing point-wise relationships $R$ among geometric neighbors. We then employ the GFT loss ${L}_{GFT}$~\cite{yang2023geometry} to measure the difference between $R_{base}$ and $R_{novel}$ on all anchors:
\begin{equation}
    \begin{aligned}
        &{L}_{GFT}=\frac{1}{Z}\sum^Z_{a=1}{||{R}_{novel}-R_{base}||}^2. \\
    \end{aligned}
\end{equation}
By minimizing this loss function, we not only transfer the structural and semantic knowledge contained in the $M_{base}$ to the $M_{novel}$ but also enhance the expression capability of $E_{novel}$ that enriches the significance of point representation. 

In terms of model structure, $M_{base}$ consists of an encoder $E_{base}$ and a classifier $Y_{base}$. For $M_{novel}$, instead of directly expanding the classifier, RDL retains the old structure consisting of the base model and introduces a residual branch to learn the residuals between the old and new classes. Hence, $M_{novel}$ consists of an encoder $E_{novel}$, a classifier $Y_{novel}$ and a residual branch $Y_{res}$. More intuitively, for instance, $Y_{novel}$ outputs the logits for base classes $\{c_1, c_2, \dots, c_n\}$, while $Y_{res}$ learns the residuals between the base and target models for $\{c_1, c_2, \dots, c_{n+1}\}$. The predictions from $Y_{novel}$ are padded with zero to match the size of $Y_{res}$, and the final output is obtained by summing $Y_{novel}$ and $Y_{res}$. This mechanism ensures that $Y_{novel}$ retains base class knowledge, while $Y_{res}$ focuses on adapting to novel classes. This process is formalized in the following equation:
\begin{equation}
    \hat{Y}_{novel}(F) = \text{pad}(Y_{novel}(F)) + Y_{res}(F).
\end{equation}
Additionally, Figure~\ref{fig:rdl}(a) visually illustrates this workflow, including the interactions between $Y_{novel}$, $Y_{res}$, and the final output.

To distill previous knowledge from the base model into the novel model, we exploit the predicted probability output by $Y_{base}$ to guide the training of $Y_{novel}$, which provides a more flexible and adaptive way to update the model's knowledge. The RDL loss function is formulated as:
\begin{equation}
    \begin{aligned}
        {L}_{RDL}=||Y_{novel}(E_{novel}(P_n))-Y_{base}(E_{base}(P_n))||^2.
    \end{aligned}
\end{equation}
This way allows for a more gradual and nuanced update of the model's parameters and enables the model to adjust its decision boundaries more smoothly.

As illustrated in Figure~\ref{fig:rdl} (b), to prevent the continuous increase in model complexity caused by structural expansion, we execute structural reparameterization to losslessly integrate the residual branch information into the main branch:
\begin{equation}
    \begin{aligned}
        \hat{Y}_n(F; \theta_{n}+ \theta_{r}) = \hat{Y}_n(F;\hat{\theta}_n + 0) = \hat{Y}_n(F;\hat{\theta}_n),
    \end{aligned}
\end{equation}
where $\theta_{n}$ and $\theta_{r}$ represent the parameters in the novel classifier $Y_{novel}(\cdot)$ and residual classifier $Y_{res}(\cdot)$, respectively. $\hat{\theta}_n$ represents reorganized parameters. $Y_{novel}(\cdot)$ is only responsible for the current base classes while $Y_{res}(\cdot)$ is responsible for the logical residuals of both the novel classes and the base classes. To ensure the dimension consistency of $\theta_{n}$ and $\theta_{r}$, we pad $\theta_{n}$ with zero and fuse with $\theta_{r}$ through element-wise addition operation. Finally, the residual branch is removed to keep the model structure consistent with $M_{base}$ for the next update. 

\subsection{Balanced Pseudo-label Learning}
\label{bpl}
\begin{figure}[t]
	\centering
	\includegraphics[width=9cm]{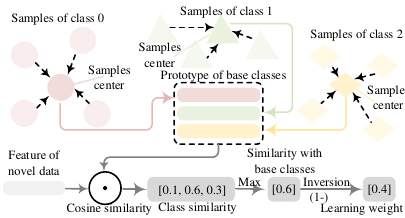}
	\caption{Illustration of our proposed BPL.}
	\label{fig:bpl}
\end{figure}

The incorporation of new information is limited by the biased number of the base and novel classes. To address such an issue, we propose a BPL strategy to eliminate the training deviation between the base and novel classes, as illustrated in Figure~\ref{fig:bpl}. Our design is based on the fact that the points with high similarity are more likely to belong to the old classes, while those with low similarity are more likely to belong to the new classes. Therefore, low-similarity samples are mainly involved in the update of the model to learn discriminative features. On the contrary, highly similar samples have a limited contribution to the update of model parameters, which promotes the learning of new class samples to the model. 

Specifically, in BPL, the prototypes serve as the representation of the base classes. We store prototypes of the base classes by computing the average embedding of all the instances belonging to each class in $C_{base}$. After projecting all new samples to the learned embedding space, we employ a standard Cosine similarity to compute the normalized cosine scores $s_{i,c}$ between them and each prototype, which is formulated as follows:
\begin{equation}
    \begin{aligned}
        s_{i,c}=Cosine(Prototype_c, E_{novel}(P_n^i)).
    \end{aligned}
\end{equation}
We utilize $s_{i,c}$ as a reference for the old-new mask generation, which addresses the model's preference towards the base classes. Points with high similarity values are more likely associated with the old classes, whereas points with low similarity values are attributed to the new classes. Furthermore, the backpropagation flow is designed to originate primarily from the new classes, updating the most discriminating positions while preserving the representation of the old classes~\cite{zhu2022self}. Therefore, we propose to generate the old-new mask as follows:
\begin{equation}
    \begin{aligned}
        m_i=1-\underset{c\in C_{base}}{\max}(s_{i,c}).
    \end{aligned}
\end{equation}

Meanwhile, to optimize the overall performance, we use the pseudo-label assigned by the old model to prevent semantic shifts. The pseudo-label generation process involves using predictions from an old model to generate labels for unlabeled data. For a given point $i$, we calculate its uncertainty $\mathcal{U}_n^i$ and further refine highly uncertain points to reduce noisy labels. The generated pseudo-label are integrated with the label of the new classes to obtain the mixed labels $\mathcal{H}_n^i$, formulated as follows:

\begin{footnotesize}  
\begin{equation}
\mathcal{H}^i_n=\left\{
\begin{aligned}
     &\mathop{\arg\max}\limits_{c\in C_{base}} Q_{b}^{i,c}   &&L^i_n= \hat{c}_{bg}, \mathop{\arg\max}\limits_{ C_{base}}Q_{b}^{i,c}\neq c_{bg}\ \text{and} \ \ \mathcal{U}_n^i\leq\gamma,
     \\&\mathop{\arg\max}\limits_{ C_{base}} Q_{b}^{k,c} &&L^i_n= \hat{c}_{bg}, \mathop{\arg\max}\limits_{ C_{base}}Q_{b}^{i,c} = c_{bg}\ \text{or} \ \ \mathcal{U}_n^i>\gamma,
     \\&L_n^i          &&L_n^i\neq \hat{c}_{bg},
     \\&\text{ignored}        &&\text{otherwise}.
\label{eq: pseudo-labels}
\end{aligned}
\right.
\end{equation}
\end{footnotesize}
\noindent In Eq.~(\ref{eq: pseudo-labels}), $L_n$ denotes the annotation from $D_{novel}$. If the pseudo-label is from $\hat{c}_{bg}$ (the background class of $C_{novel}$), and the model’s output for the current point does not match $c_{bg}$ (the background class of $C_{base}$), and the uncertainty $\mathcal{U}_n^i$ is below a threshold $\gamma$, we assign the label based on the class with the highest probability $Q^{i,c}_{b}$ from the base classes. If the pseudo-label is from $\hat{c}_{bg}$ but the model’s output either matches $c_{bg}$ or the uncertainty exceeds $\gamma$, we assign the label based on the probability $Q^{k,c}_{b}$ from the nearest neighbor. If the pseudo-label is not from $\hat{c}_{bg}$, the true label is directly used. In all other cases, the label is ignored.

Based on the generated mask $m$ and mixed labels $\mathcal{H}_n$, we design a weighting cross-entropy loss ${L}_{BPL}$ and apply it to guide the training of $M_{novel}$:
\begin{equation}
    \begin{aligned}
        {L}_{BPL}=-\sum^S_{i=1} m_i\cdot H_n^i\text{log}(Q_{n}^i).
    \end{aligned}
\end{equation}
The purpose of this design is to increase the contribution of the novel class samples during the incremental phase while encouraging the model to focus on learning the discriminate features of the novel classes. By assigning larger weights to the novel classes, we prioritize their importance during the learning process, enabling better adaptation to the new classes.

\subsection{Overall Loss}
 In the first phase, we train $M_{base}$ on points $P_b$ belonging to the base class and its labels $L_b$. We define the loss as follows:
\begin{equation}
\begin{aligned}
    {L}_{base} = - \sum^S_{i=1} L_b^i\text{log}(Y_{base}(E_{base}(P_b^i))). \\
 \end{aligned}
\end{equation} 
 In the incremental phase, the residual branch $Y_{res}$ is randomly initialized while the rest is initialized with $M_{base}$. To train $M_{novel}$, we set the total loss function as follows:
\begin{equation}
\begin{aligned}
    &{L}_{incre} = \lambda_1 {L}_{GFT} + \lambda_2 {L}_{RDL} + \lambda_3 {L}_{BPL}. \\
\end{aligned}
\end{equation}
In this paper, we empirically set $\lambda_1=1$, $\lambda_2=0.2$ and $\lambda_3=0.1$ based on careful consideration of the relative importance and impact of each loss component within the overall objective function. $\lambda_1$ follows the previous settings~\cite{yang2023geometry}. $\lambda_3$ is used to normalize $L_{BPL}$ to be of the same scale, while $\lambda_2$ is used for performance tuning, which will be discussed in Sec.~\ref{exp: hyperparameters}. After finishing the incremental phase, $M_{novel}$ is expected to accurately classify a point cloud that comes from $C_{base}\cup C_{novel}$ during the inference stage.

\section{Experiments}
\label{sec:experiment}
\subsection{Experimental Setting}
\noindent\textbf{\textit{Implementation Details}} For a fair comparison, we apply the basic setup and metric mean Intersection-over-Union (mIoU) consistent with \cite{yang2023geometry}. We adopt two paradigms $S^0$ and $S^1$ to develop $C_{base}$ and $C_{novel}$. Note that classes in $S^0$ are incrementally introduced according to the original class label order in the dataset, and classes in $S^1$ are introduced in alphabetical order. Additionally, we employ DGCNN~\cite{wang2019dynamic} as the feature extractor. Our model is trained using a batch size of 32, and the Adam optimizer~\cite{kingma2014adam} with an initial learning rate of 0.001 and weight decay of 0.0001 on both $D_{base}$ and $D_{novel}$. We train the model for 100 epochs in total. The point clouds used for training are augmented through Gaussian jitter, along with random rotations around the z-axis.\\

\begin{table*}[ht]
        \caption{Quantitative comparison of 3D class-incremental segmentation methods on the S3DIS and ScanNet datasets in $S^{0}$ split. The best results achieved by the incremental methods are highlighted in bold.}
        \label{tab:experiment:s0}
        \begin{center}
        \setlength{\tabcolsep}{9pt} 
        \renewcommand{\arraystretch}{1.02} 
        \scalebox{1}{
		\begin{tabular}{cl ccc ccc ccc}
                 \toprule[1pt]
                & &   \multicolumn{3}{c}{\textbf{$|C_{novel}|=5$}}     & \multicolumn{3}{c}{$|C_{novel}|=3$}         & \multicolumn{3}{c}{$|C_{novel}|=1$}\\
                \cline{3-11}\\
                & \multirow{1}*{Method}  & \textbf{0-7} & \textbf{8-12} & \textbf{all} & \textbf{0-9} & \textbf{10-12} & \textbf{all} & \textbf{0-11} & \textbf{12} & \textbf{all}\\
                \toprule
                \multirow{7}*{S3DIS}
                &BT &    48.54 &   -   &   -   & 46.80 &   -   &   -   & 45.00 &   -   &   -   \\
                &F\&A&   44.25 & 12.33 & 31.98 & 44.28 &  3.34 & 34.83 & 44.57 &  0.05 & 41.14 \\
                &FT &    34.96 & 30.25 & 33.15 & 28.87 & 31.56 & 29.49 & 29.44 & 29.52 & 29.45 \\
                &EWC &  39.38 & 31.07 & 36.19 & 37.13 & 37.92 & 37.31 & 36.55 & 19.94 & 35.27 \\
                &LwF &  44.55 & 35.01 & 40.88 & 43.07 & 38.34 & 41.98 & 39.94 & 35.50 & 39.60 \\
                &Yang et al.(2023a) &  48.94  & 39.56 & 45.33 & 45.15 & \textbf{45.33} & 45.19 & 44.08 & 35.69 & 43.43 \\
                &Ours &  \textbf{50.68} & \textbf{40.62} & \textbf{46.81} & \textbf{49.20} & 44.12 & \textbf{47.26} & \textbf{46.94} & \textbf{38.35} & \textbf{46.28} \\
                    \midrule[1pt]
                &JT &    50.23 & 41.74 & 46.97 & 48.62 & 41.44 & 46.97 & 47.51 & 40.41 & 46.97\\
                \midrule[1pt]
                & \multirow{1}*{Method}  & \textbf{0-14} & \textbf{15-19} & \textbf{all} & \textbf{0-16} & \textbf{17-19} & \textbf{all} & \textbf{0-18} & \textbf{19} & \textbf{all}\\
                \toprule
                \multirow{7}*{ScanNet}
                &BT &    37.37 &   -   &   -   & 34.03 &   -   &   -   & 31.57 &   -   &   -   \\
                &F\&A&   36.06 &  1.77 & 27.48 & 32.58 &  0.86 & 27.82 & 30.99 &  0.95 & 29.49 \\
                &FT &     9.39 & 13.65 & 10.45 &  8.43 & 10.98 &  8.82 &  8.02 & 10.46 &  8.14 \\
                &EWC &  17.75 & 13.22 & 16.62 & 15.70 & 11.74 & 15.11 & 15.66 &  6.76 & 15.21 \\
                &LwF &  30.38 & 13.37 & 26.13 & 26.22 & 13.88 & 24.37 & 22.15 & 12.56 & 21.67 \\
         &Yang et al.(2023a) &  \textbf{34.16} & 13.43 & 28.98 & 28.38 & 14.31 & 26.27 & 25.74 & 12.62 & 25.08 \\
                &Ours &  33.82 & \textbf{15.30} & \textbf{29.19} & \textbf{31.40} & \textbf{15.63} & \textbf{29.04} & \textbf{30.02} & \textbf{15.57} & \textbf{29.30} \\
                    \midrule[1pt]
                &JT &    38.13 & 16.63 & 32.76 & 35.46 & 17.44 & 32.76 & 33.53 & 18.08 & 32.76\\
                \bottomrule[1pt]
                \vspace{-2em}
        \end{tabular}
        }
    \end{center}
\end{table*}

\begin{table*}[!t]
        \begin{center}
        \caption{Quantitative comparison of 3D class-incremental segmentation methods on the S3DIS and ScanNet datasets in $S^{1}$ split. The best results achieved by the incremental methods are highlighted in bold.\\}
        \label{tab:experiment:s1}
        \setlength{\tabcolsep}{9pt} 
        \renewcommand{\arraystretch}{1.02} 
        \scalebox{1}{
		\begin{tabular}{cl ccc ccc ccc}
             \toprule[1pt]
            & &   \multicolumn{3}{c}{\textbf{$|C_{novel}|=5$}}     & \multicolumn{3}{c}{$|C_{novel}|=3$}         & \multicolumn{3}{c}{$|C_{novel}|=1$}\\
            \cline{3-11}\\
            & \multirow{1}*{Method}  & \textbf{0-7} & \textbf{8-12} & \textbf{all} & \textbf{0-9} & \textbf{10-12} & \textbf{all} & \textbf{0-11} & \textbf{12} & \textbf{all}\\
            \toprule
            \multirow{7}*{S3DIS}
            &BT &    37.24 &   -   &   -   & 40.73 &   -   &   -   & 45.88 &   -   &   -   \\
            &F\&A&   37.71 & 42.89 & 39.44 & 41.11 & 35.64 & 39.85 & 45.35 &  0.05 & 41.86 \\
            &FT &    10.99 & 50.67 & 26.53 & 17.83 & 54.69 & 26.34 & 23.80 &  5.74 & 22.41 \\
            &EWC &  23.19 & 54.85 & 35.36 & 29.38 & 55.53 & 35.41 & 25.60 &  9.81 & 24.39 \\
            &LwF &  32.83 & 55.19 & 41.43 & 37.69 & 54.73 & 41.62 & 32.16 & 18.26 & 31.09 \\
      &Yang et al.(2023a) & 38.17 & 55.20 & 44.72 & 39.83 & 57.59 & 43.93 & 40.33 & 19.28 & 38.71 \\
            &Ours &  \textbf{38.38} & \textbf{55.62} & \textbf{45.01} & \textbf{41.36} & \textbf{59.40} & \textbf{45.52} & \textbf{42.75} & \textbf{35.12} & \textbf{42.16} \\
                \midrule[1pt]
            &JT &    38.38 & 60.11 & 46.74 & 42.63 & 60.44 & 46.74 & 47.09 & 42.55 & 46.74\\
            \midrule[1pt]
            & \multirow{1}*{Method}  & \textbf{0-14} & \textbf{15-19} & \textbf{all} & \textbf{0-16} & \textbf{17-19} & \textbf{all} & \textbf{0-18} & \textbf{19} & \textbf{all}\\
            \toprule
            \multirow{7}*{ScanNet}
            &BT &    29.30 &   -   &   -   & 30.84 &   -   &   -   & 30.78 &   -   &   -   \\
            &F\&A&   25.25 & 18.72 & 23.62 & 26.95 &  7.37 & 24.02 & 30.41 &  0.01 & 28.89 \\
            &FT &     5.83 & 34.03 & 12.88 &  4.88 & 40.94 & 10.29 &  4.76 &  7.57 &  4.90 \\
            &EWC &  14.93 & 33.30 & 19.52 &  8.78 & 31.74 & 12.22 & 12.24 &  8.84 & 12.07 \\
            &LwF &  24.04 & 37.88 & 27.50 & 22.76 & 42.34 & 25.70 & 20.63 & 13.88 & 20.29 \\
       &Yang et al. (2023a)& \textbf{26.04} & 35.51 & 28.41 & 28.79 & 40.31 & 30.52 & 24.16 & 12.97 & 23.60 \\
            &Ours &  25.85 & \textbf{38.89} & \textbf{29.11} & \textbf{29.11} & \textbf{42.55} & \textbf{31.13} & \textbf{29.98} & \textbf{21.30} & \textbf{29.55} \\
                \midrule[1pt]
            &JT &    30.81 & 38.79 & 32.81 & 31.65 & 39.38 & 32.81 & 32.91 & 30.76 & 32.81\\
            \bottomrule[1pt]
        \end{tabular}
        }
    \end{center}
\end{table*}

\begin{table}
    \caption{Effects of different $\lambda_2$ value in setting of $|C_{novel}|=1$ on S3DIS ($S^0$ split) dataset.}
    \label{tab:lambda2}
    \centering
    \setlength{\tabcolsep}{5pt} 
    \renewcommand{\arraystretch}{1.5} 
    \scalebox{0.7}{
    \begin{tabular}{c|cccccccccc}
        \toprule[1pt]
        \textbf{Value of $\lambda_2$} & 0.1 & 0.2 & 0.3 & 0.4 & 0.5 & 0.6 & 0.7 & 0.8 & 0.9 & 1.0  \\
        \hline
        \textbf{mIoU} & 46.1 & 46.28 & 45.4 & 45.99 & 45.1 & 46.04 & 45.42 & 45.85 & 45.21 & 45.7\\
        \bottomrule[1pt]
    \end{tabular}
    }
\end{table}

\begin{table}[!t]
    \caption{Effects of different $\lambda_2$ and $\lambda_3$ in setting of $|C_{novel}|=1$ on S3DIS ($S^0$ split) dataset. The highest values are \textbf{bolded}, and the lowest values are \underline{underlined}.}
    \label{tab:lambda2_3}
    \centering
    \setlength{\tabcolsep}{8pt} 
    \renewcommand{\arraystretch}{1.5} 
    \scalebox{0.8}{
    \begin{tabular}{c|ccccc}
        \toprule[1pt]
         & $\lambda_3=$ 0.1 & $\lambda_3=$ 0.2 & $\lambda_3=$ 0.3 & $\lambda_3=$ 0.4 & $\lambda_3=$ 0.5 \\
        \hline
        $\lambda_2=$ 0.1 & 46.1 & 46.28 & 45.4 & 45.99 & \underline{45.1}\\
        $\lambda_2=$ 0.2 & 45.24 & 46.02 & 45.51 & 46.5 & 45.53\\
        $\lambda_2=$ 0.3 & 45.91 & 45.94 & 45.68 & 46.16 & 45.99\\
        $\lambda_2=$ 0.4 & 46.37 & 46.13 & 45.22 & 46.11 & 46.28\\
        $\lambda_2=$ 0.5 & 45.8 & \textbf{46.58} & 45.64 & 46.23 & 46.0\\
        \bottomrule[1pt]
    \end{tabular}
    }
\end{table}

\begin{figure*}[t]
	\centering
	\includegraphics[width=18cm]{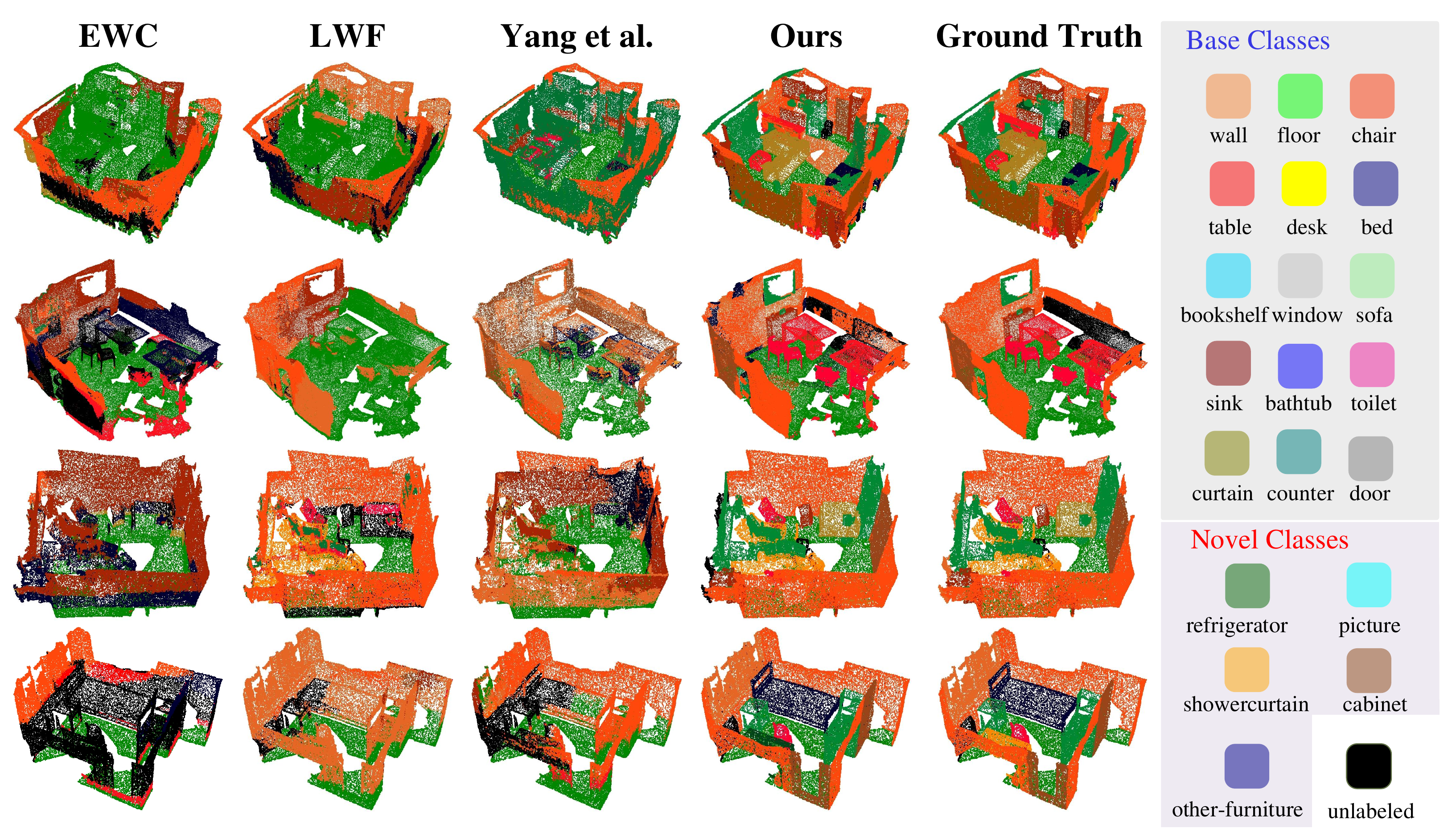}
	\caption{Visualization of Incremental Quality Comparison with EWC~\protect\cite{kirkpatrick2017overcoming}, LwF~\protect\cite{li2017learning} and \protect\cite{yang2023geometry} on $|C_{novel}|=5$ novel classes of ScanNet in order $S^0$. The results shown in black are unlabeled and do not belong to any of the base classes or the novel classes.}
	\label{fig:visual}
\end{figure*}

\begin{table*}[!t]
    \centering
    \caption{Ablation Study for multi-step training on S3DIS ($S^{0}$ split). We set the number of base classes to 8 and the number of novel classes to 5. Here, we conducted multi-step increments, where each step incremented one class, resulting in a total of five increment steps.\\}
    \label{ablation_multi-step}
    \setlength{\tabcolsep}{8pt} 
    \renewcommand{\arraystretch}{1.2} 
    \scalebox{1}{
    \begin{tabular}{c|cc|cc|cc|cc|cc}
        \toprule[1pt]
        \multirow{2}*{Class} & \multicolumn{2}{c}{\textbf{Step 1}} & \multicolumn{2}{c}{\textbf{Step 2}} & \multicolumn{2}{c}{\textbf{Step 3}} & \multicolumn{2}{c}{\textbf{Step 4}} & \multicolumn{2}{c}{\textbf{Step 5}} \\
        \cmidrule[1pt]{2-11} 
        & Yang et al. & Ours  & Yang et al.  & Ours  & Yang et al.  & Ours  & Yang et al.  & Ours  & Yang et al.  & Ours \\
        \midrule[1pt]
        0 & 88.09       & 88.03 & 85.46        & 87.33 & 85.72        & 86.81 & 85.91        & 86.71 & 88.04        & 88.40 \\
        1 & 95.63       & 96.16 & 95.66        & 96.53 & 95.76        & 96.34 & 95.37        & 96.02 & 95.83        & 96.05 \\
        2 & 73.46       & 76.54 & 71.57        & 76.04 & 72.07        & 75.96 & 65.17        & 72.58 & 65.89        & 72.81 \\
        3 &  0.00       &  0.00 &  0.00        &  0.00 &  0.00        &  0.00 &  0.00        &  0.00 &  0.00        &  0.00 \\
        4 &  7.93       &  7.50 &  0.90        &  7.03 &  0.81        &  7.03 &  0.00        &  7.24 &  0.00        &  7.27 \\
        5 & 39.70       & 49.42 & 32.01        & 48.05 & 34.79        & 45.00 & 31.28        & 41.92 & 34.48        & 43.01 \\
        6 & 22.76       & 44.29 & 19.74        & 41.82 & 11.74        & 40.11 &  6.82        & 38.84 &  8.01        & 37.70 \\
        7 & 63.06       & 60.90 & 50.07        & 58.52 & 51.78        & 57.45 & 44.29        & 58.00 & 44.38        & 59.44 \\
        \midrule[1pt]
        8 & 34.42       & 34.06 & 13.69        & 15.85 & 12.32        & 14.45 &  0.21        & 11.10 &  0.07        &  9.94 \\
        9 &   -         &   -   &  3.73        &  3.69 &  3.84        &  3.42 &  5.04        &  4.56 &  3.56        &  4.63 \\
       10 &   -         &   -   &   -          &   -   & 44.31        & 45.74 & 40.10        & 42.92 & 36.18        & 36.75 \\
       11 &   -         &   -   &   -          &   -   &   -          &   -   &  8.02        &  8.95 & 10.63        & 13.72 \\
       12 &   -         &   -   &   -          &   -   &   -          &   -   &   -          &   -   & 33.50        & 35.31 \\
        \midrule[1pt]                           
     base & 48.83       & \textbf{52.86} & 44.43        & \textbf{51.92} & 44.08        & \textbf{51.09} & 41.11        & \textbf{50.16} & 42.08        & \textbf{50.59} \\
    novel & \textbf{34.42}       & 34.06 &  8.71        &  \textbf{9.77} & 20.16        & \textbf{21.20} & 13.34        & \textbf{16.88} & 16.79        & \textbf{20.07} \\
      all & 47.23       & \textbf{50.77} & 37.28        & \textbf{43.49} & 37.56        & \textbf{42.94} & 31.85        & \textbf{39.07} & 32.35        & \textbf{38.85} \\ 
        \bottomrule[1pt]
    \end{tabular}
    }
\end{table*}

\noindent\textbf{\textit{Datasets}} We adopt two publicly available datasets: S3DIS~\cite{armeni20163d} and ScanNet~\cite{dai2017scannet} to conduct experiments, which were selected for their diversity and relevance to our problem domain and fair comparison with benchmark methods~\cite{kirkpatrick2017overcoming,li2017learning,yang2023geometry}. S3DIS comprises point clouds captured from a total of 272 rooms located in 6 different indoor areas. Each point in S3DIS contains $xyz$ coordinates and RGB information, and is manually annotated with one of the 13 predefined classes. We designate the more challenging area 5 as our validation set while using the remaining areas for training. ScanNet is an RGB-D video dataset consisting of 1,513 scans acquired from 707 indoor scenes. Each point in ScanNet is assigned to one of the 21 classes, which includes 20 semantic classes and an additional class for unannotated places. We designate the 1210 scans in the ScanNet for training, while the remaining scans are used for validation.  

\subsection{Comparison with State-of-the-arts}
We compare our method with five typical methods: Feeeze and Add (F$\&$A), Fine-Tune (FT), EWC~\cite{kirkpatrick2017overcoming}, LwF~\cite{li2017learning} and Yang et al.' method~\cite{yang2023geometry}. Among these methods, F$\&$A and FT belong to the direct adaptation methods, EWC and LwF belong to the forgetting-prevention methods, and Yang et al's method is the only open-sourced SOTA method to our best knowledge. ``BT'' corresponds to Base Training. ``JT'' represents Joint Training, where all the base + novel classes are trained together. EWC \protect\cite{kirkpatrick2017overcoming} and LwF \protect\cite{li2017learning} are traditional class-incremental methods, and \protect\cite{yang2023geometry} is state-of-the-art method currently.

The comparison results on S3DIS and ScanNet are shown in Table~\ref{tab:experiment:s0}. For F$\&$A, the performance on $C_{base}$ is maintained due to the frozen base model. However, F$\&$A faces challenges when adapting to the novel classes, leading to a subpar performance on $C_{novel}$. FT involves learning novel classes by updating the parameters of both the base feature extractor and the randomly initialized new classifier. Nonetheless, FT results in significant degradation in $C_{base}$ due to the absence of strategies to prevent forgetting. In contrast, EWC introduces the weight regularization loss to limit the modification of crucial weights from previous training, and LwF leverages the probability output by the base model to guide the model training, thereby alleviating forgetting. Additionally, \cite{yang2023geometry} explores the point-wise geometry relationship and generates uncertainty-aware pseudo-labels to prevent catastrophic forgetting and improve overall performance. Although these methods deliver good overall performance, they suffer from imbalanced performance in the base and novel classes. Our framework solved these defects well through balanced refining of the past knowledge and learning residuals. It can be seen that our BRDL achieved the best results on both the S3DIS and ScanNet datasets. Especially, the advantage of our method is obvious when $|C_{novel}|=1$. Compared to \cite{yang2023geometry}, our method improved the mIoU of novel classes by 15.34\% on S3DIS and by 8.33\% on ScanNet in $S^1$ order. A qualitative comparison is shown in Figure~\ref{fig:visual}, which demonstrates that our BRDL achieves the best performance with an outstanding balance capability compared to other methods.

\begin{table}[!t]
    \caption{Ablation study for learning strategies in $|C_{novel}|=3$ setting of S3DIS ($S^0$ \& $S^1$ split.)}
    \label{ablation-wo}
    \centering
    \setlength{\tabcolsep}{4pt} 
    \renewcommand{\arraystretch}{1.2} 
    
    \begin{tabular}{cc|ccc|ccc}
        \toprule[1pt]
          \multirow{2}*{\textbf{RDL}} & \multirow{2}*{\textbf{BPL}} &  \multicolumn{3}{c}{\textbf{$S^0$}}  & \multicolumn{3}{c}{\textbf{$S^1$}} \\
           \cmidrule(lr){3-5} \cmidrule(lr){6-8}
                    &            &\textbf{0-9}   & \textbf{10-12} & \textbf{all} & \textbf{0-9} & \textbf{10-12} & \textbf{all} \\
        \midrule[1pt] 
        \XSolidBrush&\XSolidBrush& 45.15 & 45.33 & 45.19 & 39.83 & 57.59 & 43.93 \\
         \Checkmark &\XSolidBrush& 46.12 & 40.76 & 44.90 & 41.35 & 57.43 & 45.11 \\
        \XSolidBrush& \Checkmark & 47.44 & 44.80 & 46.83 & 40.46 & 60.21 & 45.02 \\
        \Checkmark  & \Checkmark & 49.20 & 44.12 & 47.26 & 41.36 & 59.40 & 45.52 \\
        \bottomrule[1pt]
        \vspace{-2em}
    \end{tabular}
    
\end{table}

\noindent\textbf{\textit{Robustness to Class Order.}}\
We conduct experiments to demonstrate the robustness of BRDL on the above datasets in a different order (i.e. $S^0$ and $S^1$). The experimental results, as presented in Table~\ref{tab:experiment:s1}, unveils a compelling finding: for existing incremental methods, the alteration of class order does significantly affect the performance. The deviation in results indicates that the arrangement of class labels plays a crucial role in determining the success of such methods. In contrast, JT, where all categories are trained simultaneously, renders a negligible degradation in different class orders. This is because the inherent nature of JT, where the entire dataset is considered holistically, allowing for comprehensive learning across all classes, regardless of their order. Furthermore, BRDL exhibits excellent results remarkably close to those achieved through JT, even under varying class orders. This noteworthy phenomenon demonstrates the robustness and effectiveness of our approach. The above results indicate that our BRDL can adapt to different learning scenarios, making it a reliable and versatile solution for incremental learning tasks.

\subsection{Discussion of Hyperparameters}
\label{exp: hyperparameters}
The sensitivity analysis results, as shown in Table~\ref{tab:lambda2} and \ref{tab:lambda2_3}, reveal the impact of varying hyperparameters $\lambda_2$ and $\lambda_3$ on the model's performance, measured in mIoU. As shown in Table~\ref{tab:lambda2}, the model achieves its best performance when $\lambda_2$ is set to 0.2. This finding highlights that $\lambda_2$ plays a critical role in balancing the associated loss components. Moderate values like 0.2 strike a balance between regularization and generalization, whereas extreme values either overemphasize regularization ($\lambda_2$ $\geq$ 0.6) or under-regularize ($\lambda_2$ $\leq$ 0.1), leading to suboptimal model performance.

Additionally, the joint sensitivity analysis of $\lambda_2$ and $\lambda_3$, visualized in Table~\ref{tab:lambda2_3}, illustrates how the interplay between these parameters affects the model. The highest performance (mIoU = 46.58) is observed at $\lambda_2$ = 0.2 and $\lambda_3$ = 0.5, suggesting that a lower $\lambda_2$ complements moderate $\lambda_3$ values. This nuanced relationship emphasizes that while $\lambda_2$ exerts a stronger influence on overall performance, $\lambda_3$ introduces finer adjustments that can either enhance or dampen performance depending on its combination with $\lambda_2$. The analysis also shows that the model's performance is robust across a range of parameter values, with minimal fluctuation. Specifically, the performance at the highest value (46.58) and the lowest value (45.1) differs by minor, indicating that the model is not highly sensitive to small changes in these parameters. This stability highlights the robustness of the model's performance. Overall, the analysis underscores the importance of careful parameter tuning to achieve optimal results. Moderate and balanced values of $\lambda_2$ and $\lambda_3$, such as $\lambda_2$ = 0.2 and $\lambda_3$ = 0.1, are consistently effective in maximizing performance.

\subsection{Ablation Study}
To comprehensively explore BRDL, we conducted ablation studies on multi-step incremental training, learning strategies, and network initialization. The detailed explanations are as follows.

\subsubsection{Multi-step Incremental Training.}
To explore the impact of long-term learning, we conduct an ablation study to evaluate the performance of BRDL in the context of multi-step incremental training. We present the results in Table~\ref{ablation_multi-step}. The overall performance experiences a decline due to the inherent difficulty in learning novel classes. As mentioned in \cite{yang2023geometry}, the model struggles with forgetting old classes, semantic shifts in known and unknown future classes, causing confusion in recognizing previously learned classes. Despite these formidable obstacles, BRDL not only excels in preserving old knowledge but also demonstrates remarkable proficiency in acquiring new information related to the novel classes. This ability to strike a balance between retaining past knowledge and assimilating new information underscores the robustness and adaptability of BRDL in class-incremental learning scenarios. 
\begin{table}[t]
    \caption{Performance for difference initialization methods in $|C_{novel}|=3$ setting of S3DIS ($S^0$ \& $S^1$ split).}
    \label{ablation-init}
    \centering
    \setlength{\tabcolsep}{2pt} 
    \renewcommand{\arraystretch}{1.2} 
    \scalebox{1}{
        \begin{tabular}{c|ccc|ccc}
        \toprule[1pt]
             & \multicolumn{3}{c}{\textbf{$S^0$}} & \multicolumn{3}{c}{\textbf{$S^1$}} \\
        \cmidrule(lr){2-4} \cmidrule(lr){5-7}
        \multirow{-2}*{\textbf{Method}} & \textbf{0-9} & \textbf{10-12} & \textbf{all} & \textbf{0-9} & \textbf{10-12} & \textbf{all} \\
        \midrule[1pt]
            Completely random         & 46.49 & 39.27 & 44.82 & 41.79 & 59.91 & 45.97 \\   
            $Y_{res}$-zero                & 48.57 & 42.54 & 47.18 & 43.44 & 59.36 & 47.11 \\
            $Y_{res}$-random              & 49.20 & 44.12 & 47.26 & 41.36 & 59.40 & 45.52 \\
        \bottomrule[1pt]
        \end{tabular}
    }
\end{table}

\subsubsection{Learning Strategies.}
We conducted an ablation study in $|C_{novel}|=3$ setting of S3DIS in different orders, i.e., $\ S^{0}$ \& $S^{1}$, to gain insights into the individual contributions of BRDL's strategy. Table~\ref{ablation-wo} shows that the removal of any single strategy leads to a noticeable degradation in performance. As expected, RDL serves a crucial role in preventing catastrophic forgetting. When RDL is removed, the model's ability to protect old knowledge decreases, thus reducing the model performance on the old classes. Additionally, BPL contributes more to facilitate the acquisition of knowledge related to the new classes. When BPL is removed, the model's plasticity is damaged and performance on novel classes is severely declined. Although both individual strategies can help improve overall performance, the collaborative interplay between RDL and BPL culminates in the attainment of significant performance since each strategy augments the other's strengths and compensates for its limitations.

\subsubsection{Network Initialization.}
We perform a network initialization experiment to explore the impact of different network initialization strategies on the effectiveness of BRDL during training. We adopt three different strategies: completely random initialization, $Y_{res}$-random initialization, and $Y_{res}$-zero initialization. The result presented in Table~\ref{ablation-init} indicates that compared to the initialization of $Y_{res}\mbox{-}$random, the zero initialization may affect the effectiveness of the model during the training, potentially hindering its ability. Furthermore, the completely random initialization may produce significant differences in effectiveness compared to the other two methods due to insufficient training epochs. These results provide further insight into the implementation of BRDL and highlight the importance of selecting an appropriate initialization when using this technique.

\begin{table}[ht]
\centering
\caption{Comparison of computational cost between BRDL and the baseline (\protect\cite{yang2023geometry}). The comparison is conducted in S3DIS dataset ($|C_{novel}|=1, S^0$) during incremental phase.}
\scalebox{1}{
\begin{tabular}{l|c|c}
\toprule
Methods & Yang et al. (2023a) & Ours \\
\midrule
Training Time & 142 s/epoch & 144 s/epoch  \\
\hline
Memory Usage  & 1608 KB & 1614 KB \\
\bottomrule
\end{tabular}
}
\label{tab: complexity}
\end{table}

\subsection{Analysis of Computational Cost}
The comparison in Table~\ref{tab: complexity} illustrates that the computational overhead introduced by our BRDL framework is small and manageable. Specifically, the results show that our method requires 1614 KB of memory, slightly higher than the baseline, which uses 1608 KB bytes. Additionally, the training time per epoch for our method is 144 seconds, compared to 142 seconds for the baseline method. These differences arise primarily from the additional fully-connected layer and zero-padding operation incorporated into our framework. Despite these minor overheads, the performance improvements (measured in mIoU) are substantial, with our method achieving 46.94/38.35/46.28 compared to 44.08/35.69/43.43 for the baseline. This demonstrates that the tradeoff in memory usage and training time is justified, highlighting the efficiency and practicality of BRDL.

\section{Discussion}
While the BRDL framework has demonstrated promising results in addressing class-incremental learning (CIL) for 3D point clouds, there are specific conditions where its effectiveness might be limited. This section outlines potential failure modes and challenges in real-world applications.

\noindent \textbf{Class Imbalance and Small Data Volumes}
A common challenge in class-incremental learning is handling significant imbalances between old and new class data. While BRDL’s balanced pseudo-label learning (BPL) strategy addresses model bias toward base classes, extreme data imbalances could still affect the model’s ability to learn underrepresented new classes. This issue may arise in scenarios where novel classes appear infrequently or are significantly smaller in volume than older classes, causing the model to prioritize previously learned information.

\noindent \textbf{Scalability Challenges in Large Datasets}
As the scale of 3D point cloud datasets increases, the computational requirements of BRDL also grow. The need to continually distill knowledge and adapt the network to new classes could result in significant memory and processing overhead, particularly in large-scale applications like urban modeling, autonomous navigation, or industrial robotics. Without careful management of computational resources, BRDL may face scalability issues, limiting its practical application in these fields.

\noindent\textbf{Application to Real-Time Systems}
Adapting BRDL for real-time systems presents an important avenue for future research. Real-time incremental learning is critical in scenarios like autonomous vehicles and robotics, where models must adapt quickly to new environments while maintaining high accuracy. To achieve this, optimizing BRDL’s computational efficiency will be essential. Techniques such as model pruning, lightweight architectures, and early stopping during training could be explored to reduce latency and memory usage, enabling real-time deployment without compromising performance.

Overall, while BRDL provides an effective solution to many challenges in class-incremental learning for 3D point clouds, addressing these future directions will further improve the framework’s robustness and adaptability. By focusing on issues related to data quality, scalability, and real-time performance, BRDL can be refined to meet the demands of a broader range of applications, ensuring its success in both research and real-world deployment.

\section{Conclusion}
In this paper, we introduce a novel framework called Balanced Residual Distillation Learning (BRDL) for class-incremental semantic segmentation on 3D point clouds. We propose a residual distillation learning strategy to transfer and refine the knowledge previously acquired, addressing the challenge of catastrophic forgetting. This approach preserves valuable information while accommodating new learning. Additionally, we present a balanced pseudo-label learning strategy to mitigate feature confusion and reduce class bias toward base classes. Extensive experimental results demonstrate that BRDL significantly improves accuracy and generalization capabilities for both base and novel classes. Our comprehensive evaluation confirms the effectiveness and reliability of the proposed framework. We believe BRDL opens new avenues for future research in the field of class-incremental learning.

\bibliography{mybibfile}
\end{document}